# Balancing exploration and exploitation phases in whale optimization algorithm: an insightful and empirical analysis

Aram M. Ahmed[1,2], Tarik A. Rashid[3], Bryar A. Hassan[4], Jaffer Majidpour[5], Kaniaw A. Noori[6] Chnoor Maheadeen Rahman[4] Mohmad Hussein Abdalla[5], Shko M. Qader[6,7], Noor Tayfor[7] Naufel B Mohammed[2]

[1]Department of Information Technology, College of Science and Technology, University of Human Development, Sulaimani, Iraq

[2]Department of Information Technology, Kurdistan Institution for Strategic Studies and Scientific Research, Sulaimani, 46001, Iraq

[3]Department of Computer Science and Engineering, School of Science and Engineering, University of Kurdistan Hewler, Erbil, Iraq

[4]Department of Computer Science, College of Science, Charmo University, Chamchamal 46023, Sulaimani, Iraq

[5]Department of Computer Science, University of Raparin, Sulaimani, Iraq

[6]Database Technology Department, Technical College of Informatics, Sulaimani Polytechnic University, Sulaimani, Iraq

[7]Department of Computer Science, Kurdistan Technical Institute, Sulaimani, Iraq

## Abstract:

Agents of any metaheuristic algorithms are moving in two modes, namely exploration and exploitation. Obtaining robust results in any algorithm is strongly dependent on how to balance between these two modes. Whale optimization algorithm as a robust and well recognized metaheuristic algorithm in the literature, has proposed a novel scheme to achieve this balance. It has also shown superior results on a wide range of applications. Moreover, in the previous chapter, an equitable and fair performance evaluation of the algorithm was provided. However, to this point, only comparison of the final results is considered, which does not explain how these results are obtained. Therefore, this chapter attempts to empirically analyze the WOA algorithm in terms of the local and global search capabilities i.e. the ratio of exploration and exploitation phases. To achieve this objective, the dimension-wise diversity measurement is employed, which, at various stages of the optimization process, statistically evaluates the population's convergence and diversity.

## Keywords: Whale optimization algorithm, WOA, exploration and exploitation, dimension-wise diversity measurement, empirical Analysis.

## 1. Introduction

The method of choosing the best answer among many possible ones for a given problem is called optimization. The size of the search space for numerous real-world issues is a major challenge with this procedure because it makes it impossible to check every answer in a fair amount of time. Stochastic techniques that are created to address these kinds of optimization issues include algorithms that draw inspiration from nature. They often fuse deterministic and randomized procedures to produce a variety of options and then, these options are iteratively compared and analyzed until an acceptable answer is found. Stochastic algorithms can be divided into types that are single solution-based and population-based [1]. In the first one, only one agent searches the domain to find the best solution. Simulated annealing algorithm [2] is an example of this type; in contrast, population-based types like particle swarm optimization (PSO) [3], involve many agents working and cooperating with one another in a decentralized way while seeking for the optimum solution. This later type is also known as swarm intelligence. Generally, these individuals conduct their searches in two ways: global search and local search, which are also known as diversification and intensification or exploration and exploitation phases respectively. Exploration entails venturing abroad to look for fresh territories on a global scale, whereas exploitation entails concentrating on previously explored areas in order to find superior answers. Too much of global search (Over-exploration) of an algorithm will result in agent diversification and broadening, and it is rare that near-optimal solutions will be found. Too much of local search (Over-exploitation) of an algorithm, on the other hand, increases the likelihood of trapping into local optima and failing to locate the near-global optimum. As a result, there is a trade-off issue with how the two phases are balanced and maintaining a good balance is crucial in any metaheuristic algorithm [1]. Therefore, it is essential to monitor the algorithm in terms of diversification and intensification in order to actually examine the search techniques that affect these two abilities.

There are enormous tactics and methods of metaheuristic algorithms to enforce diversification and intensification in the literature [4]. They can be classified according to selection based, fitness based, subpopulation based, replacement based, hybrid based, etc. Selection techniques that encourage selecting notable agent from the existing swarm are frequently used in population-based search methods this notable agent, which is also known as global best will either incorporate on the following cycle of the search process or it will be used in some solution updating method [5]. The agents who have the best-found answers among all candidate solutions, for instance, are guaranteed to survive for the following generation in greedy selection techniques, which is known to hasten convergence toward promising solutions. Additionally, when a single greedy selection mechanism is utilized in algorithms like Deferential Evolution [6], new solutions are only chosen if they enhance the original answer or solutions. Because it pushes an initial (diverse) population to improve independently from its beginning point, this selection technique has the potential to increase the exploration-exploitation ratio of solutions [7]. Furthermore, in some algorithms such as CSO [8] the roulette wheel method is employed, and it occasionally provides the opportunity for choosing undesirable solutions as well. This will help enhance the global search ability of the algorithm. Alternatively, some search algorithms accept any new solution regardless of its quality and do not use any sort of selection technique at all. While alternative search mechanisms must be used to balance the diversification and intensification rate, these approaches for conducting searches do not necessitate the use of well-known agents. In order to improve the exploitation capabilities of algorithms like PSO [9] or DCSO [1], search techniques that make use of "attraction operators" are taken into consideration. These methods aim to enhance a swarm by either directing them in the direction of the location of agents within the swarm who appear to be "excellent" or in the direction of the location of the finest solutions currently discovered by the algorithm. The architecture of the optimization process and search strategy determines how individuals are selected as attractors and how other individuals are drawn to these attractors. Individuals are predisposed to be attracted not just to the global best solution at a given iteration of the search process, but also to the best solution(s) recorded by each particle as the search process progresses, in the case of the Particle Swarm Optimization algorithm, for example (personal best solution) [10]. In addition, Changes in swarm number, replicate, removal, infusion methods, access to outer stores, or moving from a subpopulation into another can all help achieve the balance [4]. For example, in [11 and 12] two algorithms are hybridized, and each algorithm has its own sup-population; whatever method performs better, its sub-population will increase at the expense of the other one. Then, once every while they exchange their best solutions to achieve a better balance.

AS a robust metaheuristic algorithm in the literature, the whale optimization algorithm [13], has suggested a creative plan to attain this equilibrium between exploration and exploitations. Additionally, it has produced better outcomes in a variety of applications [14]. It is crucial to examine both the behavior of each agent of WOA and the behavior of the swarm all together during the optimization process. Therefore, an insightful and empirical analyses of WOA is the main motivation behind this work.

The rest of this chapter is structured as follows: section 2 discusses exploration-exploitation Tradeoffs in in WOA. Section 3 describes the mathematical method, which is called dimension-wise diversity measurement, to compute the ratio of exploration and exploitation phases of WOA. Section 4 presents the experimental setting used in the evaluation process. Section 5 shows and discusses the results. Finally, section 6 presents the conclusion of this study.

## 2. Exploration-Exploitation Tradeoffs in in WOA

The exceptional hunting strategy of the humpback whales, which is called bubble-net feeding method is source inspiration for the WOA algorithm. During the optimization process, the best agent of each iteration is defined as the prey and the other agents in the swarm update their locations accordingly [13]. Therefore, the algorithm uses the encircling prey, Bubble-net attacking and searching for prey methods for the foraging of humpback whales, which are mathematically formulated in equations 1 to 8.

$$\vec{D} = |\vec{C} \cdot \vec{X^*}(t) - \vec{X}(t)| \tag{1}$$

$$\vec{X}(t+1) = \vec{X^*}(t) - \vec{A} \cdot \vec{D} \tag{2}$$

$$\vec{A} = 2\vec{a} \cdot \vec{r_1} - \vec{a} \tag{3}$$

$$\vec{C} = 2 \cdot \vec{r_2} \tag{4}$$

$$\vec{a} = 2\left(1 - \frac{t}{T}\right) \tag{5}$$

$$\vec{X}(t+1) = \begin{cases} \vec{X^*}(t) - \vec{A} \cdot \vec{D} & \text{if } p < 0.5 \\ D' \cdot e^{bl} \cdot \cos(2\pi l) + \vec{X^*}(t) & \text{if } p \geq 0.5 \end{cases} \tag{6}$$

$$\vec{D} = |\vec{C} \cdot \vec{X}_{\text{rand}} - \vec{X}| \tag{7}$$

$$\vec{X}(t+1) = \vec{X}_{rand} - \vec{A} \cdot \vec{D} \qquad (8)$$

Equation 1 and 2 formulates the position update for the search agents according to the prey, where $\vec{X}^*$ is the global best or prey and $\vec{X}$ is the current position of an agent. Equations 3 to 5 define $\vec{A}, \vec{C}, a$ parameters, which are used to tune and control the three above-mentioned methods.

The algorithm uses the Bubble-net attacking method as an exploitation mechanism, which is composed of two attacking approaches, namely Shrinking encircling and Spiral updating. Equation 6 represents their mathematical formulation. The first approach is accomplished by linearly lessening the value of $\vec{a}$ from 2 to 0 during the optimization process. Furthermore, this value lowers the fluctuation ranges of $\vec{A}$ as well. In addition, $r_1$ and $r_2$ are random numbers between 0 to 1, which gives an oscillation sense to $\vec{A}$ and $\vec{C}$ parameters. The second approach is accomplished by a spiral formula, which, replicates the humpback whales' helix-shaped motion. this equation considers the Euclidean distance between the individuals and the prey, which are specified through uses $D', b, l$ parameters. The algorithm makes the assumption that there is a 50% chance that the spiral model or the declining encircling model will be chosen to update the location of whales. However, in either case the algorithm attempts to gradually push the agents to move towards the global best during the optimization process. Hence the algorithm avoids the premature convergence and escapes the possible local optima.

As to the exploration capabilities, the algorithm uses same encircling mechanism, which is controlled by $\vec{A}$ parameter. In this manner, if the value of $\vec{A}$ parameter was not in the range of -1 to 1, the algorithm pushes the agents to migrate towards a random selected agent rather than the global best. Equations 7 and 8 provide mathematical formulas for the exploring phase [14].

## 3. dimension-wise diversity measurement

As mentioned earlier, Exploration and exploitation are the two primary search behaviors that swarm members often engage in. In the first, the agents are separating from one another and the spaces between them are growing wider. This stage is used to explore fresh territory and avoid any potential local optima traps. On the contrary, the agents are intensifying and getting closer together throughout the exploitation period. During this phase, they typically conduct local searches in their immediate area and congregate near the global optimum. Since, simply looking at the convergence graph and end results does not explain the exploration and exploitation ratio of WOA, dimension-wise diversity assessment [15] can be employed to quantitatively measure these phases and be able to do in-depth analysis. Additionally, this study replaced the mean in Eq. (12) with the median since it depicts the population's center more precisely [16, 17].

$$\text{Div}_j = \frac{1}{n} \sum_{i=1}^{n} \text{meadian}(x^j) - x_i^j;$$

$$\text{Div} = \frac{1}{D} \sum_{j=1}^{D} \text{Div}_j \qquad (3\text{-}8)$$

where median $(x^j)$ is the population-wide median of dimension j. n is the swarm size, and D is the dimension. $x_i^j$ is the dimension $j$ of individual $i$.

The average distance between each search agent's dimension $j$ and that dimension's median is used to calculate the diversity in each dimension, or $Div_j$. The mean diversity for all dimensions is then calculated in $Div$. By averaging the following, it is possible to determine an algorithm's percentage of exploration and exploitation:

$$XPL\% = \left(\frac{Div}{Div_{max}}\right) * 100$$

$$XPT\% = \left(\frac{|Div - Div_{max}|}{Div_{max}}\right) * 100$$

(3-9)

Where $Div_{max}$ is the highest diversity value discovered over the entire optimization procedure. Furthermore, the levels of exploration and exploitation, $XPL\%$ and $XPT\%$, respectively, are complementary.

## 4. Results and analysis

two sets of benchmarking functions were utilized to investigate the two very relevant factors (diversification and intensification) of the WOA algorithm. The first set includes of 23 traditional test functions which are of unimodal and multimodal types. Unimodal test functions often have a single global optimum in contrast to Multimodal test functions, which typically have several local optima. F1 to F7 are unimodal benchmarking functions that are used to benchmark the algorithms' capacity to perform global searches. Additionally, F8 through F23 are multimodal test functions that are used to assess the algorithms' capacity to do local searches. For a full explanation of unimodal and multimodal functions, see [13].

The other set is composed of 10 modern benchmarking functions, namely CEC 2019. These benchmark functions, also called composite benchmark functions, are challenging solve. They are extended, rotated, shifted, and merged forms of typical test functions. The comprehensive description of these benchmark functions can be found in [18].

The WOA algorithm is tested 30 times independently for each benchmark function. 30 search agents searched during the period of 500 iterations for each run. The algorithm's parameter settings are set to default, and nothing was altered. The algorithm is widely used in the literature and has been applied on enormous applications. Therefore, this chapter does not include the numerical results of the mentioned test functions. It rather sufficed with calculating the ratio or percentage of exploration and exploitation for each benchmark function merely. Table 1 and 2 presents the results for the dimension-wise diversity measurement. As it can be seen, the exploration and exploitation percentages of the algorithm are very close to each other. For the traditional benchmark functions, the average of exploration and exploitation are 51.06606522% and 48.93393% respectively. Similarly, for the CEC 2019 test functions they are 50.8883% and 49.11215%. Therefore, it can be concluded that the algorithm performs the global and local search in a balanced manner.

Table 1: exploration and exploitation ratio of WOA for the classical test functions

| Classical Test Functions | Percentage of exploration and exploitation |
|---|---|
| F1 | Exploration: 45.7196, Exploitation: 54.2804 |
| F2 | Exploration: 51.7746, Exploitation: 48.2254 |
| F3 | Exploration: 58.7007, Exploitation: 41.2993 |
| F4 | Exploration: 56.2336, Exploitation: 43.7664 |
| F5 | Exploration: 46.647, Exploitation: 53.353 |
| F6 | Exploration: 49.7779, Exploitation: 50.2221 |
| F7 | Exploration: 58.3987, Exploitation: 41.6013 |
| F8 | Exploration: 41.6489, Exploitation: 58.3511 |
| F9 | Exploration: 49.8063, Exploitation: 50.1937 |
| F10 | Exploration: 56.1418, Exploitation: 43.8582 |
| F11 | Exploration: 52.3485, Exploitation: 47.6515 |
| F12 | Exploration: 56.5455, Exploitation: 43.4545 |
| F13 | Exploration: 44.8052, Exploitation: 55.1948 |
| F14 | Exploration: 60.9302, Exploitation: 39.0698 |
| F15 | Exploration: 46.94, Exploitation: 53.06 |
| F16 | Exploration: 50.239, Exploitation: 49.761 |
| F17 | Exploration: 62.4855, Exploitation: 37.5145 |
| F18 | Exploration: 54.5377, Exploitation: 45.4623 |
| F19 | Exploration: 45.8668, Exploitation: 54.1332 |
| F20 | Exploration: 47.7914, Exploitation: 52.2086 |
| F21 | Exploration: 46.624, Exploitation: 53.376 |
| F22 | Exploration: 44.968, Exploitation: 55.032 |
| F23 | Exploration: 45.5886, Exploitation: 54.4114 |

Table 2: exploration and exploitation ratio of WOA for CEC 2019 test functions

| CEC2019 Test Functions | Percentage of exploration and exploitation |
|---|---|
| Cec01 | Exploration: 50.8883, Exploitation: 49.1117 |
| Cec02 | Exploration: 53.8929, Exploitation: 46.1071 |
| Cec03 | Exploration: 61.2, Exploitation: 38.8 |
| Cec04 | Exploration: 57.9113, Exploitation: 42.0887 |
| Cec05 | Exploration: 59.2973, Exploitation: 40.7027 |
| Cec06 | Exploration: 56.2303, Exploitation: 43.7697 |
| Cec07 | Exploration: 59.7024, Exploitation: 40.2976 |
| Cec08 | Exploration: 59.7618, Exploitation: 40.2382 |
| Cec09 | Exploration: 55.4782, Exploitation: 44.5218 |
| Cec10 | Exploration: 61.5279, Exploitation: 38.4721 |

In addition, figure 1 and 2 show the convergence curve as well as the graphical representation of exploration and exploitation capabilities of the WOA algorithm for some of the classical and modern benchmark functions. As it can be clearly seen from the figures, the algorithm usually performs high global search at the beginning, which generally ranges from 60% to 80%. Then, around iteration 300, this phenomenon reverses and the algorithm lean to decrease the diversification and increase the intensification until it reaches end of the iteration, where the exploration and exploitation phases are around 20% and 80% respectively. However, this is not the case for all the test functions. As shown in test function 12 of figure 1 and test function cec10 of figure 2, the algorithm does not tend to increase the exploitation phase and it remains around 40%.

There are many factors and elements that play important roles in an algorithm to produce robust results such as the formation of the algorithm, parameter tuning, the connection and communication between the agents, etc. The degree of diversification and intensification can also be an effective factor. As mentioned earlier, this degree in the case of WOA algorithm is close to 50% to 50%, which might reveal some of the mystery behind its success.

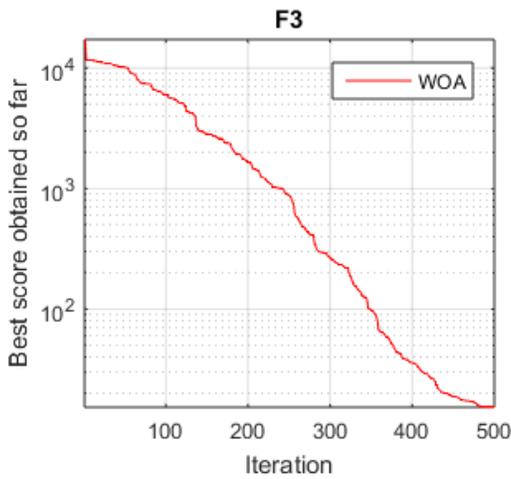
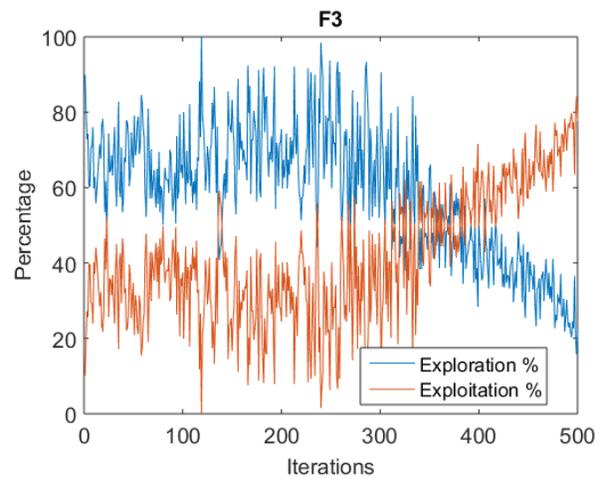
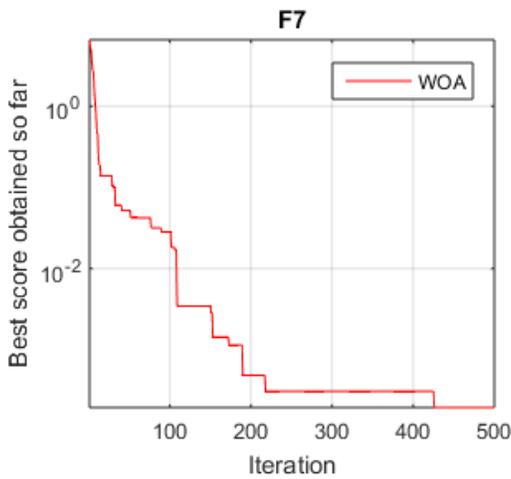
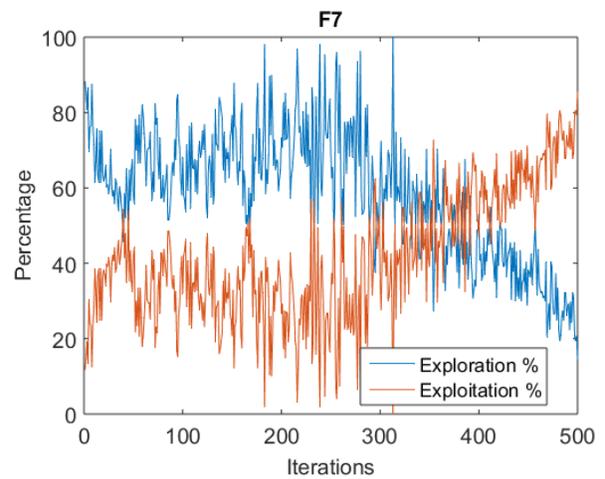

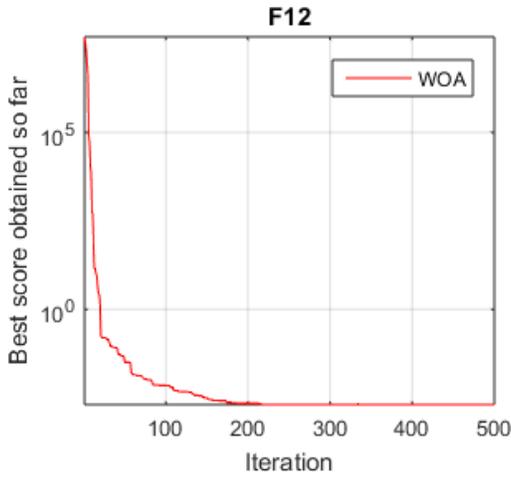
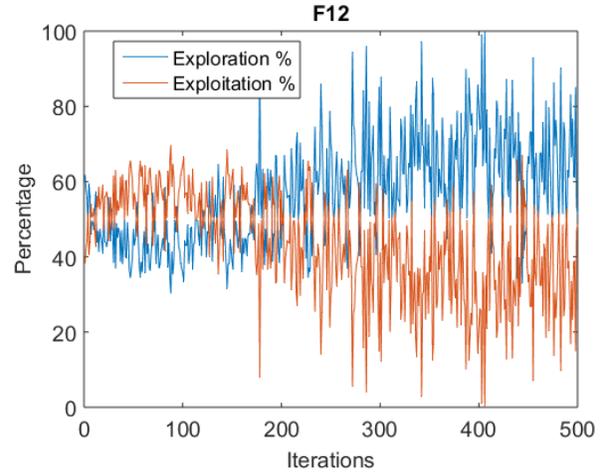
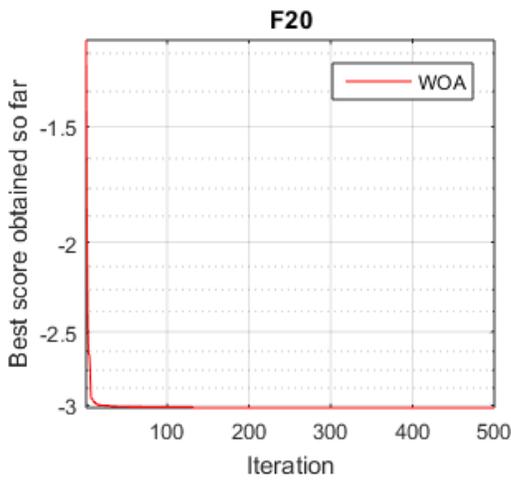
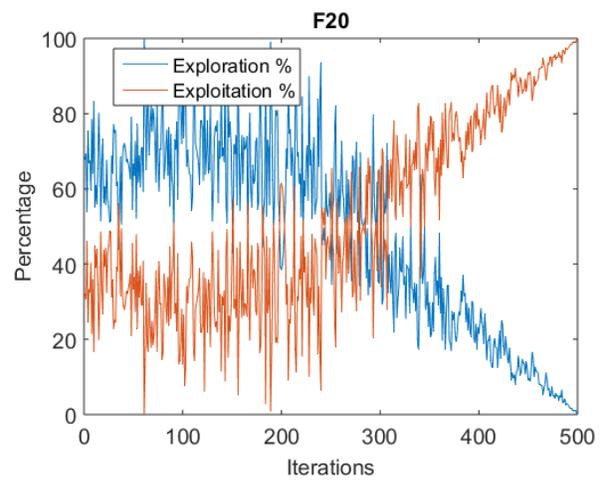

Figure 1: Convergence curve and visual representation of exploration and exploitation phases of the WOA algorithm for some of the classical test functions.

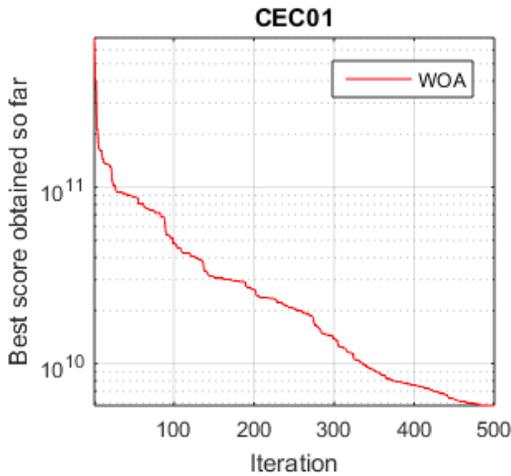
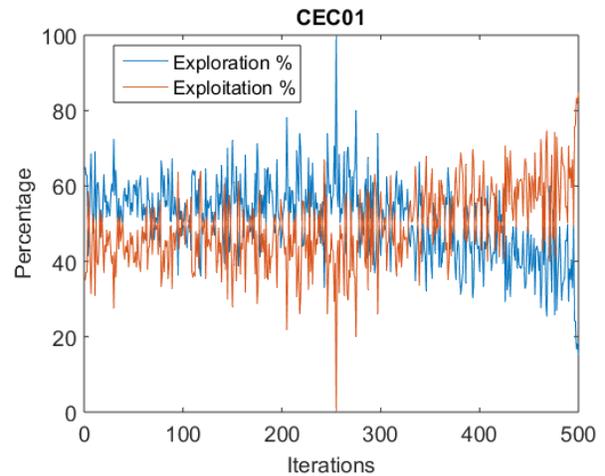

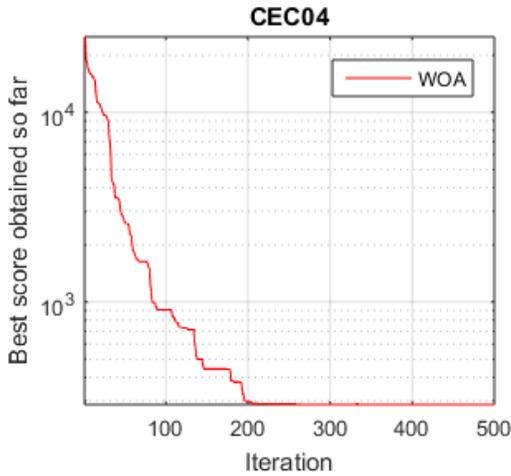
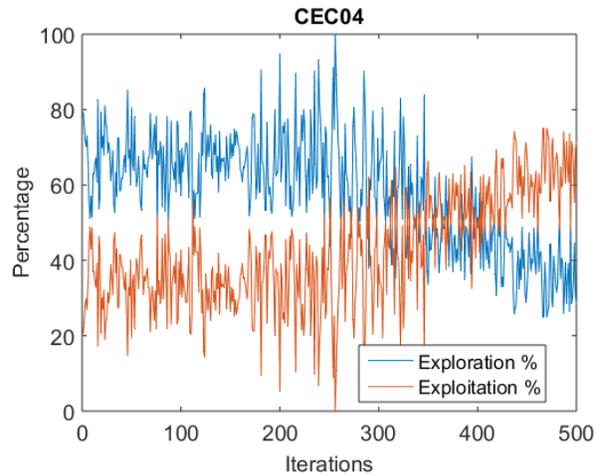
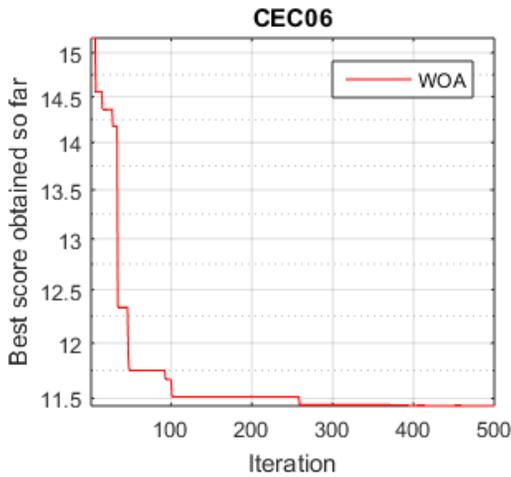
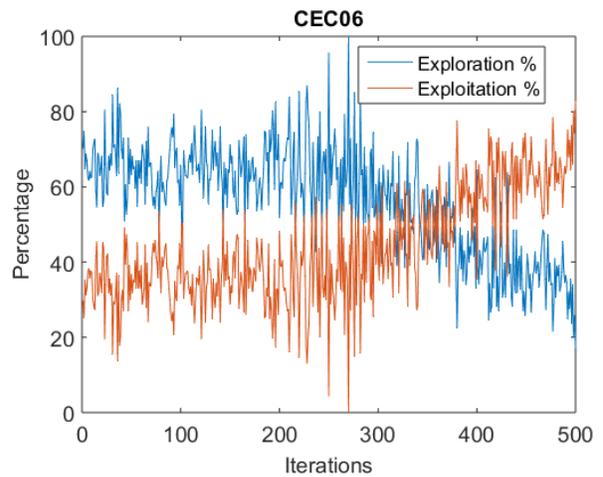
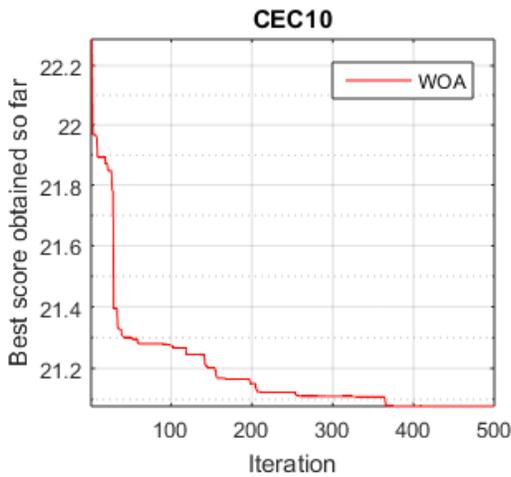
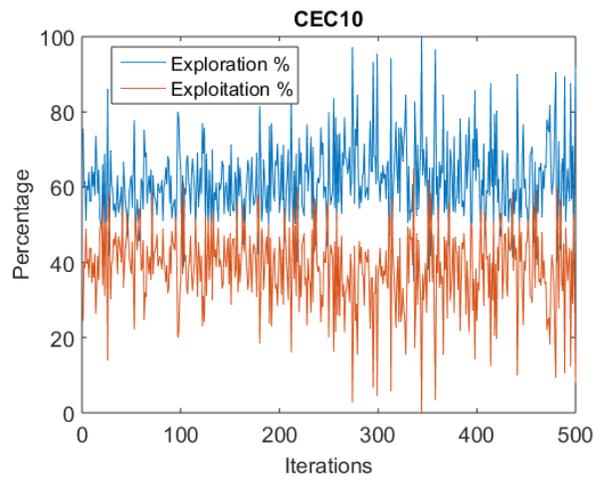

Figure 2: Convergence curve and visual representation of exploration and exploitation phases of the WOA algorithm for some of the CEC 2019 test functions.

## Summary:

This chapter attempted to empirically investigate the WOA algorithm. To achieve this, the dimension-wise diversity measurement was used to quantitatively assess the convergence and diversity population of the algorithm in different periods of the optimization process. In the experiment, two sets of benchmark functions, which were composed of 23 traditional and 10 modern benchmark functions (CEC2019), were employed. The optimization results show that the algorithm's ratio of exploration to exploitation were quite similar to one another i.e., it was close to 50% by 50%. This means that the algorithm performs the global and local searches in an even distribution and balanced manner. Moreover, the visual representation of the achieved results reveals that the algorithm possesses a great exploration rate in the beginning of the optimization process but then It gradually shifts in the direction of the local search.